\documentclass[conference]{IEEEtran}
\IEEEoverridecommandlockouts

\usepackage{cite}
\usepackage{amsmath,amssymb,amsfonts}
\usepackage{graphicx}
\usepackage{booktabs}
\usepackage{url}
\usepackage{hyperref}
\usepackage{pgfplots}
\pgfplotsset{compat=1.17}
\hypersetup{colorlinks=true,allcolors=blue}

\begin{document}

\title{Failure Modes of Always-On Inter-Cluster Repulsion\\
in Replay-Based Continual Learning}

\author{
\IEEEauthorblockN{Md Hasibul Amin}
\IEEEauthorblockA{\textit{Independent Researcher}\\
hasibulamin.cse@gmail.com}
\and
\IEEEauthorblockN{Tamzid Tanvi Alam}
\IEEEauthorblockA{\textit{Independent Researcher}\\
tamzid.t.alam@gmail.com}
}

\maketitle

\begin{abstract}
Feature-space objectives are often added to replay-based continual
learning systems with the expectation that better geometric separation
will improve retention. We study a preliminary form of
\emph{Cluster-Aware Replay} (CAR) that combines a class-balanced replay
memory with an always-active inter-cluster repulsion term (ICF). On
five-task
Split CIFAR-10 with a ResNet-18 backbone, the highest observed mean in
a six-value sensitivity sweep reaches
$22.5\pm1.4$\% final average accuracy
over three seeds, compared with
$23.1\pm2.5$\%
for replay alone. ICF without replay reaches only
$19.2\pm0.1$\%.
All tested repulsion weights produce final accuracies between
$20.1$\% and
$22.5$\%, and the detailed
configuration exhibits $89.2\pm1.5$ percentage
points of average forgetting. An instrumented rerun shows that the
weighted repulsion contribution remains near $-0.13$ after
cross-entropy has fallen close to zero, so the total objective becomes
negative while old-task accuracy collapses. Importantly, the normalized
distance objective is mathematically bounded; the failure is therefore
better described as \emph{non-saturating, always-on repulsion} rather
than an unbounded loss. These negative results show that geometric
separation is not automatically complementary to replay and motivate
margin-gated objectives whose gradients deactivate once sufficient
separation has been reached.
\end{abstract}


\section{Introduction}

Continual learning studies how a model can acquire a sequence of tasks
without erasing knowledge needed for earlier tasks
\cite{robins1995catastrophic,french1999catastrophic,de2021continual}.
Replay methods address this problem by interleaving a small memory of
old examples with current-task data and remain strong practical
baselines \cite{rebuffi2017icarl,chaudhry2018efficient,buzzega2020dark}.
A separate line of work shapes representations using prototypes,
centroids, or contrastive objectives
\cite{mai2021supervised,liu2023centroid,gu2023preserving}.

The intuition behind centroid repulsion is simple: if current features
are pushed away from previous-class centroids, classes may remain
geometrically distinct. That intuition does not guarantee a useful
continual-learning objective. A repulsion term can remain active after
classes are already separated, conflict with classification, or add no
benefit beyond replay.

This paper reports a controlled negative result for an early
Cluster-Aware Replay formulation. The study asks three questions:
(1) does always-on inter-cluster repulsion improve replay;
(2) how sensitive is the result to its weight; and
(3) what optimization behavior accompanies failure?

Our contributions are:
\begin{itemize}
    \item a three-seed evaluation and six-value sensitivity sweep showing
    that the proposed repulsion term does not improve a small replay
    baseline on Split CIFAR-10;
    \item a correction to the original terminology: with normalized
    features and centroids, the objective is bounded but remains
    non-saturating because it has no margin;
    \item an instrumented optimization trace and implementation checks
    that connect negative total-objective values with severe retention
    failure, while explicitly avoiding an unsupported claim about
    gradient dominance.
\end{itemize}

This study concerns only the preliminary always-active repulsion
objective; later margin-gated formulations and their interactions with
replay, distillation, and inference readout are outside the scope of
this paper.

\section{Related Work}

\textbf{Regularization and distillation.}
EWC and Synaptic Intelligence constrain parameter changes associated
with earlier tasks \cite{kirkpatrick2017overcoming,zenke2017continual}.
Knowledge distillation preserves outputs of a previous model
\cite{hinton2015distilling}. These methods reduce explicit storage but
can restrict plasticity.

\textbf{Replay.}
iCaRL stores class exemplars and uses prototype-based inference
\cite{rebuffi2017icarl}. GEM constrains updates using episodic memory
\cite{lopez2017gradient}, while A-GEM provides a more efficient
approximation to this constraint \cite{chaudhry2018efficient}. DER-style
methods further demonstrate the practical strength of replay-based
continual learning \cite{buzzega2020dark}.

\textbf{Feature-space structure.}
Supervised contrastive replay and centroid-distance methods explicitly
organize the representation space
\cite{mai2021supervised,liu2023centroid}. Backward feature projection
preserves separability across feature transformations
\cite{gu2023preserving}. The present study differs by analyzing a
failed repulsion objective rather than claiming a competitive method.

\section{Always-On Inter-Cluster Repulsion}

Let $f_\theta(x)$ be the feature extractor and
$\hat f_\theta(x)=f_\theta(x)/\|f_\theta(x)\|_2$. After completing
task $k$, we compute a centroid for each newly observed class
$c\in\mathcal C_k$ using all of its current-task training examples:
\begin{equation}
\tilde\mu_c =
\frac{1}{|\mathcal D_{k,c}|}
\sum_{x\in\mathcal D_{k,c}}\hat f_\theta(x),
\qquad
\mu_c=\frac{\tilde\mu_c}{\|\tilde\mu_c\|_2},
\label{eq:centroid}
\end{equation}
where $\mathcal D_{k,c}$ denotes the training examples of class $c$
in task $k$. The resulting centroid is stored for use during
subsequent tasks. Centroids computed after earlier tasks are not
refreshed as the feature extractor changes.

For a training batch $B$ and the set of previously seen classes
$\mathcal C_{<k}$, the studied inter-cluster repulsion objective
(denoted ICF, retaining the acronym of the original formulation's
Inter-Cluster Fitness term) maximizes average distance to previous
centroids:
\begin{equation}
\mathcal L_{\mathrm{ICF}}
=
-\frac{1}{|B|\,|\mathcal C_{<k}|}
\sum_{x_i\in B}
\sum_{c\in\mathcal C_{<k}}
\left\|\hat f_\theta(x_i)-\mu_c\right\|_2.
\label{eq:icf}
\end{equation}
The complete objective is
\begin{equation}
\mathcal L =
\mathcal L_{\mathrm{CE}}+
\lambda\mathcal L_{\mathrm{ICF}},
\label{eq:total}
\end{equation}
where cross-entropy is computed on the mixed current/replay training
stream. Because the inner sum in Eq.~\eqref{eq:icf} runs over the
entire mixed batch, replayed examples of previous classes are
themselves repelled from their own class centroids, placing the ICF
term in direct conflict with the classification objective on exactly
the samples replay is meant to protect.

\subsection{Why ``unbounded'' is inaccurate}

Because both arguments of each distance are unit-normalized,
\[
0\leq \|\hat f_\theta(x)-\mu_c\|_2\leq 2,
\]
and therefore $-2\leq\mathcal L_{\mathrm{ICF}}\leq0$. The loss is
bounded for every task. Its weakness is instead that it is
\emph{always active}: no margin stops repulsion after a feature is
already sufficiently far from an old centroid. The optimization target
therefore continues rewarding increased distance throughout training.

\section{Experimental Protocol}

We use Split CIFAR-10 with five tasks of two classes in the fixed order
$[0,1],[2,3],[4,5],[6,7],[8,9]$. The backbone is ResNet-18 trained from
scratch with a $3\times3$, stride-1 first convolution and no initial
max-pooling. Training uses Adam with learning rate $10^{-3}$, batch
size 32, and 20 epochs per task. The replay memory stores 20 randomly
selected examples per class.

We evaluate seeds $\{42,123,456\}$. All ``$\pm$'' values are
population standard deviations across the three per-seed scalar
metrics. We report final average accuracy and average forgetting:
\begin{equation}
F =
\frac{1}{T-1}
\sum_{i=1}^{T-1}
\left(
\max_{t\in\{i,\ldots,T\}}a_i^{(t)}-a_i^{(T)}
\right).
\end{equation}

The sweep $\lambda\in\{0.1,1,5,10,20,50\}$ is exploratory and is
reported in full. We use $\lambda=0.1$ for detailed analysis because it
has the highest observed mean in this sweep, not as an unbiased
test-selected estimate of generalization.

A separate instrumented seed-42 rerun verified that the memory contained
exactly 20 examples per observed class, centroid norms were 1.0 after
normalization, and batch labels were drawn only from classes seen so far.
It also logged cross-entropy and the weighted ICF contribution
separately.

\section{Results}

\subsection{Sequential accuracy}

Table~\ref{tab:matrix} shows the complete three-seed accuracy matrix
for $\lambda=0.1$. Earlier-task performance largely disappears after
the final task. Final average accuracy is
$22.5\pm1.4$\%, and average forgetting is
$89.2\pm1.5$ points.

\begin{table*}[t]
\centering
\caption{Per-task accuracy (\%) after each sequential task for
$\lambda=0.1$. Mean $\pm$ population standard deviation over three
seeds.}
\label{tab:matrix}
\small
\begin{tabular}{lrrrrrr}
\toprule
After training & T1 & T2 & T3 & T4 & T5 & Avg. seen\\
\midrule
$T_1$ & $98.1\pm0.2$ & -- & -- & -- & -- & 98.1 \\
$T_2$ & $21.1\pm4.1$ & $87.1\pm0.6$ & -- & -- & -- & 54.1 \\
$T_3$ & $5.9\pm0.3$ & $0.3\pm0.2$ & $91.4\pm0.6$ & -- & -- & 32.5 \\
$T_4$ & $8.0\pm3.2$ & $0.6\pm0.3$ & $1.4\pm0.4$ & $96.5\pm0.2$ & -- & 26.6 \\
$T_5$ & $0.3\pm0.2$ & $0.7\pm0.1$ & $6.7\pm7.5$ & $8.5\pm2.5$ & $96.2\pm0.2$ & 22.5 \\
\bottomrule
\end{tabular}
\end{table*}

\subsection{Sensitivity to the repulsion weight}

Table~\ref{tab:sweep} reports population standard deviations computed
across the three per-seed final average-accuracy values.

\begin{table}[t]
\centering
\caption{Final average accuracy (\%) across the ICF-weight sweep.}
\label{tab:sweep}
\begin{tabular}{cc}
\toprule
$\lambda$ & Accuracy\\
\midrule
0.1 & $\mathbf{22.5\pm1.4}$ \\
1 & $21.9\pm2.4$ \\
5 & $22.1\pm0.3$ \\
10 & $21.9\pm1.8$ \\
20 & $20.3\pm0.9$ \\
50 & $20.1\pm0.5$ \\
\bottomrule
\end{tabular}
\end{table}

\begin{figure}[t]
\centering
\begin{tikzpicture}
\begin{axis}[
    width=\columnwidth,
    height=0.70\columnwidth,
    xlabel={Repulsion weight $\lambda$},
    ylabel={Final average accuracy (\%)},
    xmode=log,
    log basis x={10},
    xmin=0.08, xmax=65,
    ymin=18.5, ymax=24,
    xtick={0.1,1,5,10,20,50},
    xticklabels={0.1,1,5,10,20,50},
    grid=both,
]
\addplot[black, thick, mark=*, mark size=1.5pt] coordinates {
    (0.1,22.500)
    (1,21.933)
    (5,22.127)
    (10,21.867)
    (20,20.293)
    (50,20.067)
};
\end{axis}
\end{tikzpicture}
\caption{The always-on repulsion objective does not produce a useful
accuracy regime in the tested sweep.}
\label{fig:sweep}
\end{figure}

Increasing $\lambda$ does not rescue retention. The best observed mean
occurs at the smallest tested weight, while the largest weights approach
the fine-tuning lower bound.

\subsection{Component ablation}

\begin{table}[t]
\centering
\caption{Ablation results on Split CIFAR-10. Accuracy is final average
accuracy over five tasks.}
\label{tab:ablation}
\small
\begin{tabular}{lc}
\toprule
Configuration & Accuracy\\
\midrule
Fine-tuning (no replay, $\lambda=0$) & $19.0\pm0.1$ \\
Replay only ($\lambda=0$) & $23.1\pm2.5$ \\
ICF only (no replay, $\lambda=0.1$) & $19.2\pm0.1$ \\
Replay + ICF ($\lambda=0.1$) & $\mathbf{22.5\pm1.4}$ \\
\bottomrule
\end{tabular}
\end{table}

Replay alone reaches
$23.1\pm2.5$\%,
while adding ICF reaches
$22.5\pm1.4$\%. With only three seeds,
the $0.6$-point difference should not be interpreted as a statistically
resolved degradation; the supported conclusion is that ICF provides no
observed improvement over replay. ICF without replay is numerically
close to the fine-tuning result.

Two observations reinforce this conclusion within the evaluated regime.
First, ICF without replay reaches $19.2\pm0.1$\%, compared with
$19.0\pm0.1$\% for fine-tuning, providing no observable standalone
improvement under this protocol. Second, none of the six tested ICF
weights produces a mean accuracy above the replay-only result of
$23.1\pm2.5$\%. These results do not establish that always-on repulsion
fails under every replay regime, but they show that it provides no
measured benefit in the configuration studied here.

\subsection{Objective-value imbalance}

Figure~\ref{fig:components} shows Task 4 from the separately
instrumented seed-42 rerun at $\lambda=0.1$. Cross-entropy falls from
$1.58$ to $0.0026$, whereas the weighted ICF contribution remains near
$-0.13$. Consequently, the total objective becomes negative from epoch
5 onward.

\begin{figure}[!b]
\centering
\begin{tikzpicture}
\begin{axis}[
    width=\columnwidth,
    height=0.72\columnwidth,
    xlabel={Epoch},
    ylabel={Objective contribution},
    xtick={1,5,10,15,20},
    grid=both,
    legend style={font=\scriptsize,at={(0.5,1.02)},anchor=south,legend columns=3},
]
\addplot[black, solid, thick, mark=*, mark size=1.5pt] coordinates {
    (1,1.5837) (5,0.0701) (10,0.0133) (15,0.0065) (20,0.0026)
};
\addlegendentry{CE}
\addplot[black, dashed, thick, mark=square*, mark size=1.5pt] coordinates {
    (1,-0.1289) (5,-0.1325) (10,-0.1329) (15,-0.1346) (20,-0.1358)
};
\addlegendentry{$\lambda \mathcal{L}_{\mathrm{ICF}}$}
\addplot[black, dotted, thick, mark=triangle*, mark size=2pt] coordinates {
    (1,1.4549) (5,-0.0625) (10,-0.1196) (15,-0.1282) (20,-0.1332)
};
\addlegendentry{Total}
\end{axis}
\end{tikzpicture}
\caption{Task-4 objective components in an instrumented seed-42 rerun.
The plot demonstrates scalar objective imbalance; it does not directly
measure component-gradient magnitudes.}
\label{fig:components}
\end{figure}

A negative total objective is not itself an optimization error, and
loss magnitudes do not prove that one component dominates parameter
updates. Establishing gradient conflict would require gradient norms and
cosine similarities. The defensible observation is narrower: the
non-saturating repulsion contribution remains substantial after
classification loss becomes very small, and this behavior coincides
with near-complete forgetting.

\section{Limitations}

This is a preliminary negative study with one dataset, one backbone,
one class order, and a small memory. The absolute accuracy regime is
also low: our replay baseline operates well below the operating points
reported for larger-memory or more heavily tuned experience-replay
systems on this benchmark. We therefore restrict our claim to this
regime: the component ablation (Table~\ref{tab:ablation}) shows no
standalone benefit over fine-tuning and no tested weight exceeding
replay-only, and we make no claim about stronger or larger-memory
replay configurations. The $\lambda$ sweep uses final test accuracy and
must therefore be viewed as exploratory. The instrumented objective
trace is a separate stochastic rerun rather than the exact checkpoint
underlying the averaged matrix. We measure objective values, not
gradient interactions.

\textbf{Centroid maintenance.}
Centroids are computed from the full class-specific training subset
when a class is first observed and are not refreshed after later
changes to the feature extractor. Consequently, centroid staleness may
contribute to the observed behavior. The study therefore diagnoses the
complete preliminary implementation---always-active repulsion together
with its original centroid-maintenance strategy---rather than
isolating the effect of non-saturation independently of centroid
drift.

Evaluating the later margin-gated formulation
would change the research question from diagnosing the original
objective to comparing revised methods, and is therefore left to
separate work.

\section{Conclusion}

Always-on centroid repulsion does not improve the tested replay system.
Across six weights, performance remains close to the fine-tuning and
replay-only baselines, and severe old-task forgetting persists. The
original description of the objective as ``unbounded'' was
mathematically imprecise: normalization bounds its value, but the lack
of a stopping margin keeps the repulsion active throughout training.

The broader lesson is that intuitive feature geometry does not
automatically translate into a useful continual-learning regularizer.
Auxiliary objectives should be evaluated through component ablations,
correct per-seed statistics, and direct optimization diagnostics before
system-level gains are attributed to them.

\section*{Data and Code Availability}

Code and reproduction scripts are available at
\url{https://github.com/AminHasibul/Continual-Learning-UsingInterClusterDistance}.
The arXiv source package accompanying this revision includes the
verified aggregate values used in every table.

\bibliographystyle{IEEEtran}
\bibliography{refs}

\end{document}